%% file: acl.tex
\pdfoutput=1

\documentclass[11pt]{article}

\usepackage[preprint]{acl/acl}

\usepackage{times}
\usepackage{latexsym}
\usepackage{xurl}        

\usepackage[T1]{fontenc}

\usepackage[utf8]{inputenc}

\usepackage{microtype}

\usepackage{inconsolata}

\usepackage{graphicx}

\input{sections_refined/custom_packages_and_defs}
\usepackage{cleveref}
\usepackage{wrapfig}

\tabcapabovefalse
\limsstartrue

\usepackage{pgfplotstable}
\usepackage{siunitx}
\usepackage{adjustbox}
\usepackage{subcaption}
\usepackage{booktabs}
\usepackage{cleveref}

\pgfplotsset{compat=1.18}
\pgfplotstableset{
  col sep=comma,
  header=true,
  string type,
  empty cells with={\textemdash} 
}

\title{Steering LLM Reasoning Through Bias-Only Adaptation}

\newcommand{\affmark}[1]{\textsuperscript{#1}}

\newlength{\authcolwd}
\newcommand{\authorbox}[1]{\makebox[\authcolwd][c]{#1}}

\author{\authorbox{Viacheslav Sinii\affmark{1}\affmark{*}} \\\And
  \authorbox{Nikita Balagansky\affmark{1,2}} \\\And
  \authorbox{Yaroslav Aksenov\affmark{1}} \\\AND
  \authorbox{Vadim Kurochkin\affmark{1,2}} \\\And
  \authorbox{Daniil Laptev\affmark{1}} \\\And
  \authorbox{Gleb Gerasimov\affmark{1,2,3}} \\\AND
  \authorbox{Alexey Gorbatovski\affmark{1,2}} \\\And
  \authorbox{Boris Shaposhnikov\affmark{1}} \\\And
  \authorbox{Daniil Gavrilov\affmark{1}}  \\
}

\author{
 \textbf{Viacheslav Sinii\textsuperscript{1}},
 \textbf{Alexey Gorbatovski\textsuperscript{1}},
 \textbf{Artem Cherepanov\textsuperscript{1,2}},
 \textbf{Boris Shaposhnikov\textsuperscript{1}},
\\
 \textbf{Nikita Balagansky\textsuperscript{1}},
 \textbf{Daniil Gavrilov\textsuperscript{1}},
\\
\\
 \textsuperscript{1}T-Tech,
 \textsuperscript{2}Central University,
\\
 \small{
   \textbf{Correspondence:} \href{mailto:v.siniy@t-tech.dev}{v.siniy@t-tech.dev}
 }
}

\begin{document}
\maketitle

\input{sections_refined/abstract}
\input{sections_refined/introduction}
\input{sections_refined/related_work}
\input{sections_refined/methodology}
\input{sections_refined/results}
\input{sections_refined/conclusion}

\input{sections_refined/limitations}
\input{sections_refined/acknowledgment}

\bibliography{custom}

\clearpage
\onecolumn
\appendix

\input{sections_refined/appendices/steering_vector_vis}
\input{sections_refined/appendices/steering_vector_implementation}
\input{sections_refined/appendices/training_objective}
\input{sections_refined/appendices/lora}
\input{sections_refined/appendices/old_experiments}
\input{sections_refined/appendices/gpt_prompt}
\input{sections_refined/appendices/training_cost_savings}
\input{sections_refined/appendices/computational_resources}
\input{sections_refined/appendices/hyperparameters}

\end{document}

%% file: sections_refined/custom_packages_and_defs.tex
\usepackage{amsfonts}
\usepackage{amsmath}
\usepackage{booktabs}
\usepackage{tabularx}
\usepackage{CJKutf8}          
\usepackage[most]{tcolorbox}  
\usepackage[cachedir=minted-cache]{minted}
\usepackage{caption}       
\usepackage{adjustbox}
\usepackage{multirow}
\usepackage{cleveref}

\usepackage{makecell}
\colorlet{softgreen}{green!50!black}

\newcommand{\gapplus}[1]{\rlap{\kern.5em\scriptsize\textcolor{softgreen}{\,(+#1)}}}
\newcommand{\gapminus}[1]{\rlap{\kern.5em\scriptsize\textcolor{red}{\,(-#1)}}}
\newcommand{\gapzero}{\rlap{\kern.5em\scriptsize\textcolor{gray}{\,(0.00)}}}

\tcbset{
  pygbox/.style={
    colback=gray!12,
    colframe=blue!75!black,
    boxrule=0.8pt,
    arc=2mm,
    left=0pt,right=0pt,top=0pt,bottom=0pt, 
    enhanced,
  }
}

\newif\iftabcapabove
\newif\iflimsstar

%% file: sections_refined/abstract.tex
\begin{abstract}
We show that training a single $d$-dimensional steering vector per layer with reinforcement learning, while freezing all base weights, matches the accuracy of fully RL-tuned reasoning models on mathematical-reasoning tasks.
On an 8 billion-parameter model this adds only $\approx 0.0016\%$ additional parameters and reproduces performance across a range of base models and mathematical-reasoning benchmarks.
These results tighten the upper bound on the parameter budget required for high-level chain-of-thought reasoning, indicating that millions of adapter weights are unnecessary.
The minimal trainable footprint reduces optimizer memory and inter-GPU communication, lowering the overall cost of fine-tuning.
Moreover, a logit-lens analysis shows that the learned vectors amplify coherent token directions, providing clearer insight into the model’s internal computations.
The code is available at \url{https://github.com/corl-team/steering-reasoning}.
\end{abstract}

%% file: sections_refined/introduction.tex
\section{Introduction}

Reasoning models have recently made striking gains by learning to produce a chain-of-thought before the final answer \citet{jaech2024openai, guo2025deepseek}. Much of this progress comes from reinforcement learning with verifiable rewards in mathematical domains, a now-standard setup for training reasoning models \citep{zeng2025simplerl, hu2025open, venhoff2025understanding}. However, training large models is costly, and, because of the amount of parameters and complex internal computations, the mechanisms induced by reasoning training remain poorly understood.

In this work, we show that training just $0.0016\%$ of parameters suffices to match the performance of a fully RL-tuned model. We train per-layer steering vectors that are added to each layer’s output while keeping all base weights fixed (see \Cref{appendix:steering_vector_vis} for visualization). Compared with LoRA \citep{hu2022lora} and BitFit \citep{zaken2021bitfit}, this approach \textbf{(i)} requires orders of magnitude fewer resources and \textbf{(ii)} isolates a much smaller, more interpretable parameter set, making it easier to see what changes during reasoning training.

We present a preliminary study showing that these vectors amplify meaningful directions in representation space, aligning with interpretable token clusters such as causality (“Because”, “However”), validation (“correctness”, “necessity”, “confirmation”), and programming-language tokens.

Taken together, these results provide a simple, resource-efficient training setup that both reduces the cost of adapting large models and simplifies the study of how reasoning training modifies pretrained models.

\input{tables/accuracies}

%% file: tables/accuracies.tex
\makeatletter
\typeout{DEBUG: vm=\ifvmode V\else H\fi, inner=\ifinner T\else F\fi}
\makeatother
\begin{table*}[t]
\centering
\begingroup
\setlength{\tabcolsep}{4pt}        
\renewcommand{\arraystretch}{1.0}
\begin{adjustbox}{max totalsize={\textwidth}{\textheight},center}
\begin{filecontents*}{tables/accuracies_std.csv}
Model,Setup,AIME25 AVG@32,AIME25 PASS@1,AIME24 AVG@32,AIME24 PASS@1,AMC23 AVG@32,AMC23 PASS@1,Avg.,MATH500,MinervaMath,OlympiadBench
Qwen2.5-1.5B,Base,0.0 $\pm$ 0.0,0.0 $\pm$ 0.0,0.0 $\pm$ 0.0,0.0 $\pm$ 0.0,0.0 $\pm$ 0.0,0.0 $\pm$ 0.0,0.3 $\pm$ 0.0,0.9 $\pm$ 0.0,0.7 $\pm$ 0.0,0.3 $\pm$ 0.0
Qwen2.5-1.5B,Steering,1.4 $\pm$ 0.0,1.1 $\pm$ 1.6,2.8 $\pm$ 0.1,4.4 $\pm$ 4.2,34.1 $\pm$ 0.1,28.3 $\pm$ 3.1,23.2 $\pm$ 0.3,57.5 $\pm$ 1.5,20.3 $\pm$ 1.8,23.3 $\pm$ 0.6
Qwen2.5-1.5B,Full-Tune,1.2 $\pm$ 0.0,0.0 $\pm$ 0.0,4.1 $\pm$ 0.2,3.3 $\pm$ 2.7,29.9 $\pm$ 0.2,24.2 $\pm$ 4.2,22.7 $\pm$ 0.4,58.4 $\pm$ 0.7,20.6 $\pm$ 1.5,22.2 $\pm$ 1.1
Qwen2.5-1.5B,SimpleRL-Zoo,,,4.2 $\pm$ 0.0,6.7 $\pm$ 0.0,,35.0 $\pm$ 0.0,,59.0 $\pm$ 0.0,20.2 $\pm$ 0.0,21.0 $\pm$ 0.0
Qwen2.5-1.5B,Open-Reasoner,,1.0 $\pm$ 0.0,,3.5 $\pm$ 0.0,,,,58.0 $\pm$ 0.0,,
Qwen2.5-7B,Base,6.7 $\pm$ 0.0,6.7 $\pm$ 0.0,13.3 $\pm$ 0.0,13.3 $\pm$ 0.0,39.2 $\pm$ 0.0,39.2 $\pm$ 0.0,24.2 $\pm$ 0.0,45.9 $\pm$ 0.0,10.2 $\pm$ 0.0,29.9 $\pm$ 0.0
Qwen2.5-7B,Steering,7.0 $\pm$ 0.3,13.3 $\pm$ 5.4,12.8 $\pm$ 0.0,13.3 $\pm$ 5.4,50.5 $\pm$ 0.1,55.0 $\pm$ 2.0,36.4 $\pm$ 0.3,74.7 $\pm$ 1.0,36.6 $\pm$ 1.1,36.6 $\pm$ 1.2
Qwen2.5-7B,Full-Tune,6.6 $\pm$ 0.2,4.4 $\pm$ 1.6,13.6 $\pm$ 0.2,14.4 $\pm$ 3.1,53.8 $\pm$ 0.1,53.3 $\pm$ 5.1,37.1 $\pm$ 0.5,77.3 $\pm$ 0.7,34.6 $\pm$ 2.3,37.2 $\pm$ 0.5
Qwen2.5-7B,Open-Reasoner,,15.6 $\pm$ 0.0,,17.9 $\pm$ 0.0,,,,81.4 $\pm$ 0.0,,
Qwen2.5-7B,Open-Reasoner (from Oat-Zero),,,,13.3 $\pm$ 0.0,,54.2 $\pm$ 0.0,,82.2 $\pm$ 0.0,31.6 $\pm$ 0.0,47.9 $\pm$ 0.0
Qwen2.5-7B,SimpleRL-Zero,,,15.6 $\pm$ 0.0,20.0 $\pm$ 0.0,,62.5 $\pm$ 0.0,,78.2 $\pm$ 0.0,38.6 $\pm$ 0.0,40.4 $\pm$ 0.0
Qwen2.5-7B,R1-Distill,,,,33.0 $\pm$ 0.0,,68.4 $\pm$ 0.0,,88.1 $\pm$ 0.0,35.9 $\pm$ 0.0,47.7 $\pm$ 0.0
Qwen2.5-14B,Base,3.3 $\pm$ 0.0,3.3 $\pm$ 0.0,6.7 $\pm$ 0.0,6.7 $\pm$ 0.0,37.5 $\pm$ 0.0,37.5 $\pm$ 0.0,26.1 $\pm$ 0.0,61.7 $\pm$ 0.0,20.3 $\pm$ 0.0,26.8 $\pm$ 0.0
Qwen2.5-14B,Steering,15.4 $\pm$ 0.3,14.4 $\pm$ 4.2,15.9 $\pm$ 0.4,20.0 $\pm$ 0.0,59.8 $\pm$ 0.4,62.5 $\pm$ 2.0,42.3 $\pm$ 0.2,80.0 $\pm$ 0.2,39.2 $\pm$ 1.7,43.5 $\pm$ 1.1
Qwen2.5-14B,Full-Tune,12.1 $\pm$ 0.6,16.7 $\pm$ 0.0,15.9 $\pm$ 0.4,16.7 $\pm$ 5.4,59.2 $\pm$ 0.2,57.5 $\pm$ 3.5,41.3 $\pm$ 0.1,80.5 $\pm$ 0.1,38.6 $\pm$ 0.3,41.3 $\pm$ 1.4
Qwen2.5-Math-1.5B,Base,3.3 $\pm$ 0.0,3.3 $\pm$ 0.0,13.3 $\pm$ 0.0,13.3 $\pm$ 0.0,27.5 $\pm$ 0.0,27.5 $\pm$ 0.0,18.1 $\pm$ 0.0,32.2 $\pm$ 0.0,9.4 $\pm$ 0.0,22.9 $\pm$ 0.0
Qwen2.5-Math-1.5B,Steering,8.1 $\pm$ 0.1,8.9 $\pm$ 1.6,12.1 $\pm$ 0.0,12.2 $\pm$ 3.1,48.8 $\pm$ 0.1,51.7 $\pm$ 3.1,33.9 $\pm$ 0.3,70.8 $\pm$ 0.8,27.5 $\pm$ 0.7,36.1 $\pm$ 0.5
Qwen2.5-Math-1.5B,Full-Tune,8.4 $\pm$ 0.1,7.8 $\pm$ 4.2,11.9 $\pm$ 0.1,14.4 $\pm$ 1.6,49.1 $\pm$ 0.2,51.7 $\pm$ 6.2,33.6 $\pm$ 0.1,70.1 $\pm$ 0.2,29.4 $\pm$ 0.9,32.8 $\pm$ 0.5
Qwen2.5-Math-1.5B,Oat-Zero,,,,20.0 $\pm$ 0.0,,53.0 $\pm$ 0.0,,74.2 $\pm$ 0.0,25.7 $\pm$ 0.0,37.6 $\pm$ 0.0
Qwen2.5-Math-7B,Base,3.3 $\pm$ 0.0,3.3 $\pm$ 0.0,16.7 $\pm$ 0.0,16.7 $\pm$ 0.0,45.8 $\pm$ 0.0,45.8 $\pm$ 0.0,24.8 $\pm$ 0.0,52.2 $\pm$ 0.0,12.3 $\pm$ 0.0,18.6 $\pm$ 0.0
Qwen2.5-Math-7B,Steering,12.6 $\pm$ 0.2,20.0 $\pm$ 4.7,24.4 $\pm$ 0.2,28.9 $\pm$ 4.2,62.5 $\pm$ 0.2,61.7 $\pm$ 5.1,43.3 $\pm$ 0.3,79.9 $\pm$ 1.2,36.0 $\pm$ 2.8,44.1 $\pm$ 0.8
Qwen2.5-Math-7B,Full-Tune,14.0 $\pm$ 0.1,14.4 $\pm$ 3.1,25.7 $\pm$ 0.1,30.0 $\pm$ 4.7,64.2 $\pm$ 0.1,64.2 $\pm$ 3.1,43.5 $\pm$ 0.4,79.3 $\pm$ 0.6,36.9 $\pm$ 2.0,41.1 $\pm$ 0.9
Qwen2.5-Math-7B,Oat-Zero,,,,43.3 $\pm$ 0.0,,62.7 $\pm$ 0.0,,80.0 $\pm$ 0.0,30.1 $\pm$ 0.0,41.0 $\pm$ 0.0
Qwen2.5-Math-7B,SimpleRL-Zero,,,24.0 $\pm$ 0.0,40.0 $\pm$ 0.0,,70.0 $\pm$ 0.0,,80.2 $\pm$ 0.0,37.5 $\pm$ 0.0,39.0 $\pm$ 0.0
LLaMa3.1-8B-It,Base,0.0 $\pm$ 0.0,0.0 $\pm$ 0.0,10.0 $\pm$ 0.0,10.0 $\pm$ 0.0,28.3 $\pm$ 0.0,28.3 $\pm$ 0.0,21.7 $\pm$ 0.0,52.3 $\pm$ 0.0,21.1 $\pm$ 0.0,18.4 $\pm$ 0.0
LLaMa3.1-8B-It,Steering,1.5 $\pm$ 0.2,1.1 $\pm$ 1.6,10.3 $\pm$ 0.2,12.2 $\pm$ 6.8,32.4 $\pm$ 0.7,35.0 $\pm$ 7.1,26.1 $\pm$ 0.6,58.7 $\pm$ 0.8,29.3 $\pm$ 1.5,24.6 $\pm$ 1.3
LLaMa3.1-8B-It,Full-Tune,1.6 $\pm$ 0.2,1.1 $\pm$ 1.6,10.8 $\pm$ 0.2,7.8 $\pm$ 1.6,35.3 $\pm$ 0.4,38.3 $\pm$ 1.2,26.1 $\pm$ 0.1,58.0 $\pm$ 0.7,29.4 $\pm$ 0.6,21.8 $\pm$ 0.8
LLaMa3.1-8B,Base,0.0 $\pm$ 0.0,0.0 $\pm$ 0.0,0.0 $\pm$ 0.0,0.0 $\pm$ 0.0,5.0 $\pm$ 0.0,5.0 $\pm$ 0.0,3.9 $\pm$ 0.0,11.2 $\pm$ 0.0,5.1 $\pm$ 0.0,2.0 $\pm$ 0.0
LLaMa3.1-8B,Steering,0.3 $\pm$ 0.0,0.0 $\pm$ 0.0,0.3 $\pm$ 0.1,3.3 $\pm$ 2.7,8.9 $\pm$ 0.1,11.7 $\pm$ 2.4,9.1 $\pm$ 0.1,25.7 $\pm$ 0.9,12.4 $\pm$ 0.3,7.1 $\pm$ 0.4
LLaMa3.1-8B,Full-Tune,0.5 $\pm$ 0.1,0.0 $\pm$ 0.0,1.6 $\pm$ 0.0,1.1 $\pm$ 1.6,9.0 $\pm$ 0.2,9.2 $\pm$ 2.4,11.5 $\pm$ 0.4,29.8 $\pm$ 1.6,18.4 $\pm$ 2.1,9.9 $\pm$ 0.8
\end{filecontents*}
\pgfplotstabletypeset[
  columns={Model,Setup,{AIME25 AVG@32},{AIME24 AVG@32},{AMC23 AVG@32},{MATH500},{MinervaMath},{OlympiadBench},{Avg.}},
  every column/.style={column type=c},
  display columns/0/.style={string type, column type={>{}l}},
  display columns/1/.style={string type, column type={>{}l}},
  columns/{AIME25 AVG@32}/.style={
    column name={AIME25},
    column type=c
  },
  columns/{AIME24 AVG@32}/.style={
    column name={AIME24},
    column type=c
  },
  columns/{AMC23 AVG@32}/.style={
    column name={AMC23},
    column type=c
  },
  every head row/.style={before row=\toprule, after row=\midrule},
  every last row/.style={after row=\bottomrule},
  every row no 4/.style={after row=\midrule},
  every row no 11/.style={after row=\midrule},
  every row no 14/.style={after row=\midrule},
  every row no 18/.style={after row=\midrule},
  every row no 23/.style={after row=\midrule},
  every row no 26/.style={after row=\midrule},
]{tables/accuracies_std.csv}
\end{adjustbox}
\endgroup

\caption{Accuracies on six mathematical-reasoning benchmarks for three variants of each model: the Base model (no training), a Fully-Tuned model, and our Steering model that trains only per-layer steering vectors while freezing all other weights. For reference we also include numbers reported by SimpleRL-Zoo, Open-Reasoner, Oat-Zero, and R1-Distill. Across models and datasets, Steering matches the performance of Fully-Tuned models.}
\label{tab:accuracies}
\end{table*}

%% file: sections_refined/related_work.tex
\section{Related Work}
\label{sec:related}

The use of steering vectors, a technique within activation engineering, provides a direct way to probe and manipulate model behavior with minimal changes to the underlying weights. Traditionally, such vectors are constructed from activation differences on contrastive prompts (e.g., positive vs. negative sentiment) and are typically interpreted as feature amplifiers rather than creators of novel behaviors \citep{turner2023st_act_eng, panickssery2023st_contr}.  They have already been used to identify and control "reasoning" behaviours \citep{venhoff2025understanding, ward2025reasoning}. Other work demonstrates that these vectors can also be trained, not merely computed, allowing for more targeted control. For example, \citet{cao2024personalized} optimized steering directions using preference data, while \citet{mack2024melbo} and \citet{engels2025mechanisms} (building on \citet{betley2025tell}) showed that training simple additive vectors in an unsupervised manner can elicit complex latent behaviors, such as reasoning and self-awareness. We apply these ideas on a scale of a real GRPO-like training, and show that their simlicity and interpretability benefits do not stand in constrast to quality of the trained models.

We implement steering vectors only by activating only layer-wise additive biases in MLPs while keeping all other model parameters frozen. This approach is aligned with BitFit \citep{zaken2021bitfit}, which tunes only bias terms and has been shown to effectively expose existing knowledge, often matching the performance of full fine-tuning on language tasks. Notably, BitFit and similar minimal-adaptation methods sometimes underperform on tasks requiring substantial generalization \citep{hu2022lora}; it remains unclear whether they suffice for complex reasoning. And still, differently from BitFit, we tune only a subset of model biases, tightening the training landscape even more. These minimal interventions stand in contrast to parameter-efficient finetuning methods such as prompt tuning and LoRA \citep{hu2022lora, li2025llms}, or full RL-based adaptation \citep{guo2025deepseek}, which actively adjust model parameters. 

Our work is partly motivated by the recent results on reasoning training only amplifying the  reasoning behaviors already present in the base models \citep{wang2025reinforcement, ye2025limo, shao2025spurious, liu2025oatzero}. If such a minimal intervention is able to recover reasoning performance, this adds as an evidence to such a claim. Our aim is to take the study further by identifying what specific parts of the model are holding these behaviours.

\begingroup
\def\clusterscaption{Clusters of tokens most aligned with the learned steering vectors, as measured by cosine similarity.}
\begin{CJK*}{UTF8}{gbsn} 
\begin{table*}[t]
  \centering
  \iftabcapabove
  \caption{\clusterscaption}
  \fi
  \begin{tabularx}{\linewidth}{p{0.05\linewidth} p{0.15\linewidth} p{0.28\linewidth} X@{}}
    \toprule
    \textbf{Layer idx} & \textbf{Top-Cluster} & \textbf{Representative tokens} & \textbf{Unifying idea} \\ \midrule
    \textbf{2} & \textbf{Source-code \& test-harness vocabulary} &
    \texttt{tostring}, \texttt{ComponentFixture}, \texttt{.SQL}, \texttt{standalone}, \texttt{-independent}, \texttt{\_fault}, \texttt{203}, \texttt{@a}, \texttt{\textbackslash Context} &
    These are the words you meet in programming projects - Angular’s ComponentFixture, SQL file extensions, "fault" flags, HTTP status 203, context objects, and helper functions like toString(). \\ 
    \midrule
    \textbf{2} & \textbf{Named entities (people \& places)} &
    \texttt{Antonio}, \texttt{Pelosi}, \texttt{Baldwin}, \texttt{Cumberland}, \texttt{Switzerland}, \texttt{Peg}, \texttt{Salv-} &
    Proper names of individuals and locations that commonly co-occur in news articles or knowledge-graph dumps. \\ 
    \midrule
    \textbf{17} & \textbf{Accuracy, validation \& logical necessity} &
    \texttt{correctness}, \texttt{correct}, \texttt{precision}, \texttt{necessity}, \texttt{possibility}, \texttt{confirmation}, \texttt{answer}, \texttt{goal}, \texttt{directly}, \texttt{derive} / \texttt{deriving} &
    These words belong to discourse about getting things right - arguments, proofs, validations, QA reports, or formal specifications. \\
    \midrule
    \textbf{30} & \textbf{Causal \& contrastive connectors} &
    \texttt{Because / because / 因为}, \texttt{Therefore / donc},
    \texttt{However / однако / jedoch}, \texttt{Given / Here}, \texttt{step} &
    Words that introduce reasons, consequences, or contrasts - typical of argumentative writing, technical explanations, or test-case descriptions. \\

    \bottomrule
  \end{tabularx}
  
  \iftabcapabove\else
  \caption{\clusterscaption}
  \fi
  \label{tab:clusters}
\end{table*}
\end{CJK*}

\endgroup

%% file: sections_refined/methodology.tex
\section{Methodology}
\subsection{Online Training}

We adopt an online reinforcement learning procedure loosely modeled on DeepSeek-R1 \citep{ahmadian2024back, guo2025deepseek}.
For each prompt $x$, we sample $N$ candidate solutions $y_1, \ldots, y_N$ from the current policy $\pi_\theta$. Each rollout $y_i$ receives a binary reward $r(x,y_i)$ based on the presence of a correct answer enclosed in a \texttt{\textbackslash boxed\{\dots\}} template. The model is trained with RLOO \citep{ahmadian2024back} objective. Other details are in \Cref{appendix:training_math}.


\subsection{Steering Vector}

We insert a learnable \textbf{steering vector} $s_\ell \in \mathbb{R}^d$ at the end of every transformer layer $\ell$ (there are $L$ layers in total). The vector is added directly to the residual stream, so its dimensionality matches the model’s hidden size $d$. All original weights remain frozen; only these $L$ steering vectors are trained. \Cref{appendix:steering_vector_vis} has a visualisation and \Cref{appendix:steering_vector_impl} contains our code implementation for clarity.

\subsection{Training and Evaluation Setup}

We experiment across multiple model families and sizes: \texttt{Qwen2.5-\{1.5,7,14\}B} \citep{qwen2.5}, \texttt{Qwen2.5-Math-\{1.5,7\}B} \citep{yang2024qwen25mathtechnicalreportmathematical}, \texttt{Llama3.1-8B}, and \texttt{Llama3.1-8B-Instruct} \cite{grattafiori2024llama}. Training uses the DeepScaleR dataset \citep{deepscaler2025} with sampling temperature $\tau=1$, a 4K context for \texttt{Qwen2.5-Math-\{1.5,7\}B}, and 8K for the other models, 128 prompts per step.

We report results on six math benchmarks: AIME24/25, AMC23, MATH500 \citep{hendrycks2021measuring}, MinervaMath \citep{lewkowycz2022solving}, and OlympiadBench \citep{he2024olympiadbench}. For MATH500, MinervaMath, and OlympiadBench we report \textsc{Pass@1}; for AIME24/25 and AMC23 we report \textsc{Avg@32} due to their smaller size. Base models decode greedily, whereas trained models use sampling with $\tau=1.0$ following \citet{zeng2025simplerl}. Evaluation context length is 4K and 32K for Qwen2.5-Math-7B and other models respectively. All metrics are averaged over three evaluation seeds. Unless noted otherwise (e.g., \Cref{tab:accuracies}), figures and tables show the mean score across the six benchmarks. When available, we include numbers from SimpleRL-Zoo \citep{zeng2025simplerl}, Oat-Zero \citep{liu2025understanding}, Open-Reasoner \citep{hu2025open}, and R1-Distill \citep{guo2025deepseek} for context. Implementations use \texttt{transformers} \citep{wolf2019huggingface}, the \texttt{vllm} inference engine \cite{kwon2023efficient}, and \texttt{Math-Verify}\footnote{\url{https://github.com/huggingface/Math-Verify}} for reward assignment.

%% file: sections_refined/results.tex
\section{Results}

\subsection{Steering Vectors are Effective for Inducing Reasoning Capabilities}


\input{tables/training_cost_savings_qwen}
\Cref{tab:accuracies} shows that steering vectors match the accuracy of fully tuned models across families and scales. The only exception is LLaMa3.1-8B, where steering recovers about $70\%$ of the full-tuning gain. See \Cref{appendix:old_experiments} for results when training on the GSM8K and MATH datasets.

Because only a fraction of parameters is optimized, the approach yields substantial savings (\Cref{tab:training_cost_savings_qwen}): the optimizer state shrinks to kilobytes and the parameter broadcast\footnote{We broadcast trainable model parameters to \texttt{vllm} inference engine after each parameter update.} time drops by nearly an hour per training epoch. Measurements for Qwen2.5-7B and LLaMa3.1-8B-It are detailed in \Cref{appendix:training_cost_savings}.

\subsection{Interpretation}

To understand what the learned steering vectors are doing inside the network, we apply the logit-lens technique \citep{nostalgebraist2020lens}.
The key idea is to "peek" into a residual stream after a specific transformer layer by converting it into a full vocabulary distribution and then reading the most likely tokens.

Let the row $u_v \in \mathbb{R}^d$ of $W_U$ correspond to token $v$.
For every token we compute the cosine similarity

$$
c_l(v)\;=\;\frac{\langle s_l,\;u_v\rangle}{\|s_l\|\,\|u_v\|}\;\in[-1,1].
$$

A large positive $c_l(v)$ means the steering vector pushes the hidden state toward token $v$; a large negative value indicates suppression. 

We collect top-$50$ tokens for each steering vector and ask GPT-o3 to translate all non-english tokens and group the subsets of tokens into explainable topics (see the prompt in Appendix~\ref{appendix:gpt_prompt}).

Table~\ref{tab:clusters} shows the representative token groups from different layers of \texttt{LLaMa3.1-8B-It} model trained on \texttt{GSM8K} dataset. 
At layer 2, the steering vector aligns with programming-style terms rather than math tokens, which is surprising given the math-oriented task. While not being from the math domain, this use of coding tokens suggests the model leverages structural parallels between programming and formal math notation. It also picks up named entities because many \texttt{GSM8K} problems use character names and places to set up word problems. 

At layer 17, the vector shifts to words about checking steps and validating results. This suggests the model uses the middle stage to verify each reasoning step before proceeding.

At layer 30, it focuses on linking words such as "because", "therefore", and "however". This indicates the final stage ties statements together to guide the answer’s flow.

Overall, the learned steering vectors appear to be highly interpretable and relevant to the reasoning task on the \texttt{GSM8K} domain.

%% file: tables/training_cost_savings_qwen.tex

\begin{table}[h]
  \vspace{-1.2em}
  \centering
  \begin{tabular}{lcc}
    \toprule
    \textbf{Metric} & \textbf{Full-Tune} & \textbf{Steering} \\ \midrule
    Number of Parameters & 14.7 B  & 245 K \\
    Optimizer Memory     & 13.8 GB & 240 KB \\
    Per-step Time        & 9.94 s & 0.11 s \\
    Overall Time & 52 m & 34 s \\ \bottomrule
  \end{tabular}
  \caption{Resource cost for Qwen2.5-14B: full fine-tuning vs.\ steering. Overall Time is across 314 steps $\approx$ 1 epoch.}
  \label{tab:training_cost_savings_qwen}
\end{table}\ignorespaces

%% file: sections_refined/conclusion.tex
\section{Conclusion}
In this paper, we have demonstrated that training lightweight steering vectors alone can recover the reasoning performance of fully‐tuned models on standard mathematical benchmarks. This result carries two important implications. First, steering vector training offers a highly parameter‐efficient and cost‐effective alternative: only a small set of layer‐wise bias terms must be learned, drastically reducing storage and communication time requirements. Second, it isolates a small, interpretable set of parameters that capture the effects of reasoning training, simplifying the study of the mechanisms of models' reasoning abilities.

%% file: sections_refined/limitations.tex
\iflimsstar
\section*{Limitations}
\else
\section{Limitations}
\fi

First, our experiments cover a narrow slice of online-training settings. Broader sweeps -- across settings, tasks, and model sizes -- would test generality and may reveal cases where steering vectors fall short of full fine-tuning.

Second, while the logit-lens provides a convenient way to inspect how steering vectors influence token logits at each layer, it does not capture the downstream transformations applied by subsequent layers. As a result, later computations may modify the initial steering signal, leading to interpretations of logit-lens itself that conflict with layer-wise observations. Applying more comprehensive interpretation techniques, such as probing classifiers, causal interventions, or circuit-level analysis, could yield deeper insights into how steering vectors shape the model’s behavior.

%% file: sections_refined/acknowledgment.tex
\section*{Acknowledgment}
\label{section:ack}

Viacheslav dedicates his contribution in this paper to his girlfriend, Marina.

%% file: sections_refined/appendices/steering_vector_vis.tex
\section{Steering Vector Visualization}
\label{appendix:steering_vector_vis}

\begin{figure}[H]
  \centering
  \includegraphics[width=\textwidth]{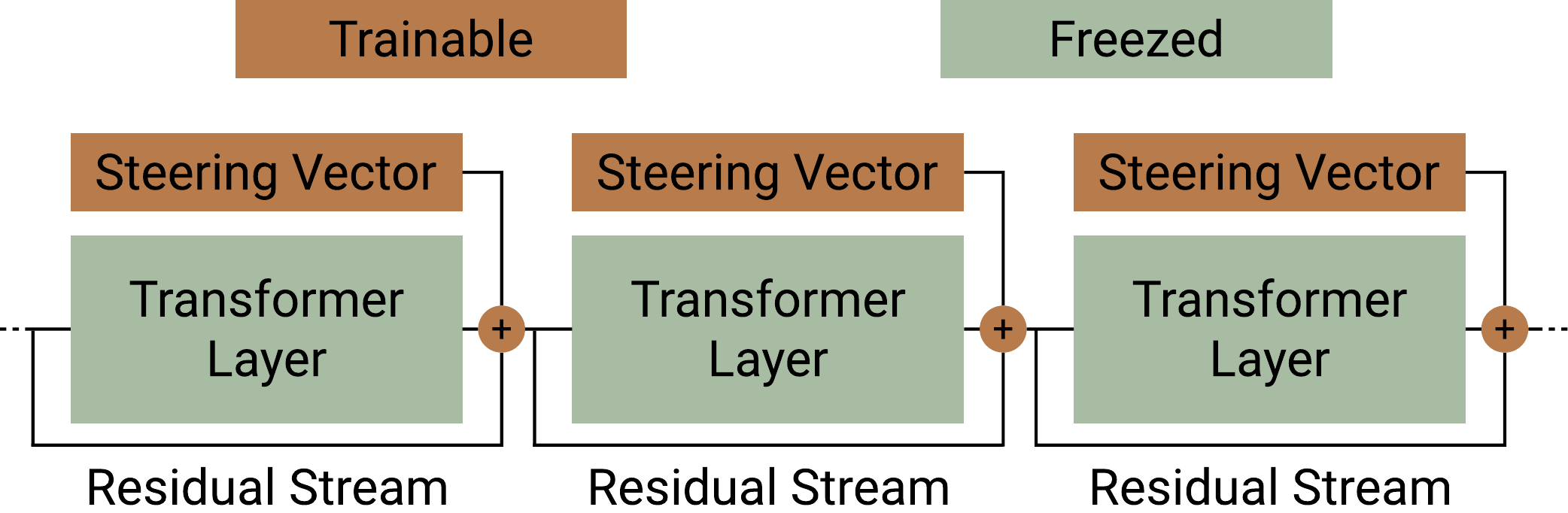}
  \caption{\textbf{Layer-wise trainable steering vectors.} All base transformer weights are frozen (blue). The only trainable parameters are one $d$-dimensional vector $v_\ell$ per layer (orange), added to the residual stream at every token position: $h_{\ell,t}\ \leftarrow\ h_{\ell,t} + v_\ell$.}
  \label{fig:steering_vector_vis}
\end{figure}

%% file: sections_refined/appendices/steering_vector_implementation.tex
\section{Steering Vector Implementation}
\label{appendix:steering_vector_impl}

\begin{tcolorbox}[pygbox,title=Steering Vector Implementation]
\begin{minted}[]{python}
class SteeringVector(nn.Module):
    def __init__(self, hidden_size: int):
        super().__init__()

        self.hidden_size = hidden_size

        self.steering_vector = nn.Parameter(
            torch.zeros(self.hidden_size).unsqueeze(0).unsqueeze(0)
        )

    def forward(self, x):
        return x + self.steering_vector
        
class TransformersQwen2DecoderLayerWithSteering(TransformersQwen2DecoderLayer):
    def __init__(self, config: Qwen2Config, layer_idx: int):
        super().__init__(config=config, layer_idx=layer_idx)

        self.steering_vector = SteeringVector(hidden_size=config.hidden_size)

        self.layer_idx = layer_idx

    def forward(self, *args, **kwargs):
        hidden_states, *rest = super().forward(*args, **kwargs)

        hidden_states = self.steering_vector(hidden_states)

        return (hidden_states, *rest)
        
\end{minted}
\end{tcolorbox}

%% file: sections_refined/appendices/training_objective.tex
\section{Training Objective}
\label{appendix:training_math}
To reduce variance, we compute a baseline $b$ as the mean reward for all rollouts associated with $x$:

$$
b = \frac{1}{N} \sum_{i=1}^N r_i, \qquad a_i = r_i - b.
$$

The parameters are updated via a policy-gradient step:

$$
\nabla_\theta J = \mathbb{E}_{x \sim D, y \sim \pi_\theta(x)}\bigl[a(x,y)\,\nabla_\theta\log\pi_\theta(y \mid x)\bigr].
$$

%% file: sections_refined/appendices/lora.tex
\section{LoRA}
\label{appendix:lora}

A limitation of steering vectors is that the same vector is added to every token position. We hypothesize that this token-independence may cap performance and partly account for the gap between the full-model baseline and steering-only training.

To let the offset be token-specific, we replace the fixed steering vector with a low-rank adaptor (LoRA) \cite{hu2022lora} applied to the MLP down-projection in every transformer layer:

$$
\begin{aligned}
\text{(fixed steering)} &\quad h' = h + s,\\[4pt]
\text{(LoRA steering)} &\quad h' = h + B\cdot A \cdot h_{\text{MLP}},
\end{aligned}
$$
where $h \in \mathbb{R}^{d}$ is the residual stream, $s \in \mathbb{R}^{d}$ is a learned constant, $h_{\text{MLP}} \in \mathbb{R}^{d_{\text{MLP}}}$ is the intermediate representation of MLP layer, and $A$ and $B$ are LoRA rank-$r$ matrices which are the only trainable components in this setup.

All experiments use LoRA rank $r = 4$, scaling factor $\alpha = 4$, and no dropout following~\citet{engels2025mechanisms}.

%% file: sections_refined/appendices/old_experiments.tex
\section{Steering Vectors on GSM8K and MATH}
\label{appendix:old_experiments}

\paragraph{Experimental Setup.}
We conduct experiments on four pretrained transformer checkpoints: \texttt{Qwen-2.5-1.5B} \citep{qwen2.5}, \texttt{Qwen-2.5-Math-1.5B} \citep{yang2024qwen25mathtechnicalreportmathematical}, \texttt{Llama-3.1-8B}, and \texttt{Llama-3.1-8B-Instruct} \cite{grattafiori2024llama}. For each model, we evaluate three training regimes: (i) full fine-tuning, (ii) training only steering vectors, and (iii) training only LoRA \citep{hu2022lora} adapters  which may be viewed as adaptive steering vectors (Appendix~\ref{appendix:lora}, \citet{mack2024melbo}). For LoRAs we use rank $4$. In the latter two cases, all other parameters are kept frozen.

We use two mathematical datasets for training and evaluation. The \texttt{GSM8K} training split contains 8,790 problems \cite{gsm8k}; for evaluation, we randomly subsample 500 items from its original split to shorten iteration time. The \texttt{MATH} corpus \cite{math} provides 12,000 training examples, and its evaluation split consists of 500 items, which we use as is. We report $\text{mean}@8 = \mathbb{E}_x[\mathbb{E}_{i=0}^8\ r(x,y)]$ win rates on each dataset. 

All experiments are implemented using the \texttt{transformers} library \citep{wolf2019huggingface} and the \texttt{vllm} inference engine \cite{kwon2023efficient}. Additional hyperparameter details are provided in Appendix~\ref{appendix:hypers}.

\input{tables/old_accuracies}
\paragraph{Result.}

\Cref{tab:perf_single} summarizes accuracy for the pretrained model, fully-tuned models, steering vectors, and LoRA adapters. As expected, RL training yields large gains on mathematical benchmarks on all setups. Steering vectors achieve similar improvements across nearly all model-dataset pairs and even exceed full fine-tuning in some cases (e.g., \texttt{Qwen2.5-1.5B} when evaluated on \texttt{MATH-500} and \texttt{LLaMa3.1-8B} when trained on \texttt{GSM8K} and evaluated on \texttt{MATH-500}, both being base models that did not undergo instruction tuning), which we attribute to the implicit regularization of updating far fewer parameters. 

If we accept the working assumption that a steering vector can only amplify features that the original network already contains and cannot create new ones, the table gives direct evidence for that view: when the base model "knows" how to solve the task, steering is usually enough to reach the same quality as full fine-tuning.

There are, however, a few setups where the steering training stays noticeably below the full training - for example, the \texttt{Qwen2.5-Math-1.5B} and \texttt{LLaMa3.1-8B} evaluated on \texttt{GSM8K}. In most cases the LoRA closes the gap completely.  Because LoRA modifies a small, learned set of rank-decomposed weight matrices rather than a single global vector, it provides finer control over what is added to the residual stream. The fact that LoRA always bridges the remaining gap shows that a more targeted, low-rank adjustment is the reason why  single steering vector cannot reach the performance of a fully-trained model.

%% file: tables/old_accuracies.tex
\begingroup

\def\perfsinglecaption{$\text{mean}@8$ accuracy for each combination of training dataset, evaluation dataset, model, and tuning setup.  
  Rows are grouped by \textbf{Train / Test} dataset pairs, and each column corresponds to a specific model variant.  
  For \textit{Steering} and \textit{LoRA} rows, the colored value in parentheses indicates the difference compared to \textit{Full-Tune} for that model - \textcolor{softgreen}{green} if better, \textcolor{red}{red} if worse.  
  This highlights how close lightweight methods can get to full fine-tuning performance, and where gaps remain.}

\begin{table*}[t]

  \centering
  
  \iftabcapabove
  \caption{\perfsinglecaption}
  \fi
  \adjustbox{max width=\linewidth}{
  \begin{tabular}{llcccc}
    \toprule
    \textbf{Train / Test} & \textbf{Setup}
      & \textbf{Qwen2.5-1.5B}
      & \textbf{Qwen2.5-Math-1.5B}
      & \textbf{Llama3.1-8B}
      & \textbf{Llama3.1-8B-It} \\
    \midrule
    \multirow{4}{*}{GSM8K / GSM8K}
      & Base      & 0.63 & 29.26 & 1.08 & 66.03 \\
      & Full-Tune & 78.91 & 86.49 & 76.49 & 87.22 \\
      & Steering  & 73.84\gapminus{5.07}
                  & 79.89\gapminus{6.60}
                  & 70.36\gapminus{6.13}
                  & 87.22\gapzero \\
      & LoRA      & 76.49\gapminus{2.42}
                  & 85.41\gapminus{1.08}
                  & 74.24\gapminus{2.25}
                  & 85.41\gapminus{1.81} \\
    \midrule
    \multirow{4}{*}{GSM8K / MATH}
      & Base      & 1.51 & 28.73 & 0.76 & 32.51 \\
      & Full-Tune & 40.78 & 65.95 & 16.86 & 46.65 \\
      & Steering  & 48.69\gapplus{7.91}
                  & 61.79\gapminus{4.16}
                  & 22.81\gapplus{5.95}
                  & 48.51\gapplus{1.86} \\
      & LoRA      & 49.32\gapplus{8.54}
                  & 64.31\gapminus{1.64}
                  & 19.28\gapplus{2.42}
                  & 49.04\gapplus{2.39} \\
    \midrule
    \multirow{4}{*}{MATH / MATH}
      & Base      & 1.51 & 28.73 & 0.76 & 32.51 \\
      & Full-Tune & 44.48 & 70.44 & 27.39 & 52.39 \\
      & Steering  & 51.39\gapplus{6.91}
                  & 65.27\gapminus{5.17}
                  & 22.05\gapminus{5.34}
                  & 50.81\gapminus{1.58} \\
      & LoRA      & 53.68\gapplus{9.20}
                  & 69.35\gapminus{1.09}
                  & 24.32\gapminus{3.07}
                  & 50.40\gapminus{1.99} \\
    \midrule
    \multirow{4}{*}{MATH / GSM8K}
      & Base      & 0.63 & 29.26 & 1.08 & 66.03 \\
      & Full-Tune & 68.57 & 82.56 & 52.12 & 85.03 \\
      & Steering  & 69.56\gapplus{0.99}
                  & 76.41\gapminus{6.15}
                  & 45.24\gapminus{6.88}
                  & 84.02\gapminus{1.01} \\
      & LoRA      & 72.53\gapplus{3.96}
                  & 81.88\gapminus{0.68}
                  & 52.52\gapplus{0.40}
                  & 85.06\gapplus{0.03} \\
    \bottomrule
  \end{tabular}}

  \iftabcapabove\else
  \caption{\perfsinglecaption}
  \fi
  \label{tab:perf_single}
\end{table*}

\endgroup

%% file: sections_refined/appendices/gpt_prompt.tex
\section{Logit Lens. GPT Prompt}
\label{appendix:gpt_prompt}
\begin{tcolorbox}[pygbox,title=GPT Prompt for Token Clustering]
\begin{minted}{text}
You will be given a list of tokens together with a score. 
You should translate all non-english tokens and suggest the main topics 
that unite the biggest subsets of tokens in the list.

<list>
\end{minted}
\end{tcolorbox}

%% file: sections_refined/appendices/training_cost_savings.tex
\section{Training Cost Savings}
\label{appendix:training_cost_savings}

\input{tables/training_cost_savings}

%% file: tables/training_cost_savings.tex

\begin{table}[H]
  \centering

  %
  %
  \begin{subtable}[t]{0.48\linewidth}
    \centering
    \caption{Number of Parameters}
    \label{tab:number-of-parameters}
    \begin{tabular}{lccc}
      \toprule
         & Qwen2.5-7B & Llama3.1-8B \\
      \midrule
      Full-Tune & 7.6 B & 8 B   \\
      Steering  & 100 K & 131 K \\
      \bottomrule
    \end{tabular}
  \end{subtable}\hfill
  \begin{subtable}[t]{0.48\linewidth}
    \centering
    \caption{Optimizer Memory}
    \label{tab:optimizer-memory}
    \begin{tabular}{lccc}
      \toprule
         & Qwen2.5-7B & Llama3.1-8B \\
      \midrule
      Full-Tune & 7.1 GB & 7.5 GB \\
      Steering  & 98 KB  & 128 KB \\
      \bottomrule
    \end{tabular}
  \end{subtable}

  \bigskip 

  %
  %
  \begin{subtable}[t]{0.48\linewidth}
    \centering
    \caption{Per-step Time}
    \label{tab:per-step-time}
    \begin{tabular}{lccc}
      \toprule
         & Qwen2.5-7B & Llama3.1-8B \\
      \midrule
      Full-Tune & 5.30 s & 5.32 s \\
      Steering  & 0.06 s & 0.07 s \\
      \bottomrule
    \end{tabular}
  \end{subtable}\hfill
  \begin{subtable}[t]{0.48\linewidth}
    \centering
    \caption{Overall Time (314 steps $\approx$ 1 epoch)}
    \label{tab:overall-time}
    \begin{tabular}{lccc}
      \toprule
         & Qwen2.5-7B & Llama3.1-8B \\
      \midrule
      Full-Tune & 27.7 m  & 27.8 m \\
      Steering  & 0.314 m & 0.36 m \\
      \bottomrule
    \end{tabular}
  \end{subtable}

  \caption{Resource‐efficiency comparison of full fine-tuning versus steering across three model sizes.}
  \label{tab:training_cost_savings}
\end{table}

%% file: sections_refined/appendices/computational_resources.tex
\section{Computational Resources}
All models were trained on $16$ H100 GPUs. \texttt{Qwen2.5-1.5B} models were trained for approx.~$9$ hours, \texttt{Qwen2.5-Math-1.5B} for approx.~$2.5$ hours, \texttt{LLaMa3.1-8B-It} for approx.~$9$ hours, \texttt{LLaMa3.1-8B-It} for approx.~$120$ hours.

%% file: sections_refined/appendices/hyperparameters.tex
\section{Hyperparameters}
\label{appendix:hypers}

\begin{table}[H]
  \centering
  \adjustbox{max width=\linewidth}{\input{tables/hypers}}
  \caption{Hyperparameter settings for each model and training setup.}
  \label{tab:hyperparams}
\end{table}

%% file: tables/hypers.tex
\begin{tabular}{@{}lllcc@{}}
\toprule
 &  &  & $\mathrm{lr}$ & $\mathrm{num\_generations}$ \\
\midrule
\multirow[t]{8}{*}{Qwen-2.5-1.5B} & \multirow[t]{4}{*}{GSM8K} & Full-Tune & $2 \times 10^{-5}$ & 64 \\
 &  & Steering & $5 \times 10^{-4}$ & 64 \\
 &  & LoRA-1 & $5 \times 10^{-4}$ & 64 \\
 &  & LoRA-4 & $5 \times 10^{-4}$ & 64 \\
\cline{2-5}
 & \multirow[t]{4}{*}{MATH} & Full-Tune & $2 \times 10^{-5}$ & 64 \\
 &  & Steering & $5 \times 10^{-4}$ & 64 \\
 &  & LoRA-1 & $5 \times 10^{-4}$ & 64 \\
 &  & LoRA-4 & $5 \times 10^{-4}$ & 64 \\
\cline{1-5} \cline{2-5}
\multirow[t]{8}{*}{Qwen-2.5-Math-1.5B} & \multirow[t]{4}{*}{GSM8K} & Full-Tune & $2 \times 10^{-5}$ & 16 \\
 &  & Steering & $1 \times 10^{-3}$ & 16 \\
 &  & LoRA-1 & $5 \times 10^{-4}$ & 16 \\
 &  & LoRA-4 & $5 \times 10^{-4}$ & 16 \\
\cline{2-5}
 & \multirow[t]{4}{*}{MATH} & Full-Tune & $2 \times 10^{-5}$ & 16 \\
 &  & Steering & $1 \times 10^{-3}$ & 16 \\
 &  & LoRA-1 & $5 \times 10^{-4}$ & 16 \\
 &  & LoRA-4 & $5 \times 10^{-4}$ & 16 \\
\cline{1-5} \cline{2-5}
\multirow[t]{8}{*}{Llama-3.1-8B} & \multirow[t]{4}{*}{GSM8K} & Full-Tune & $5 \times 10^{-6}$ & 64 \\
 &  & Steering & $5 \times 10^{-4}$ & 64 \\
 &  & LoRA-1 & $1 \times 10^{-4}$ & 64 \\
 &  & LoRA-4 & $1 \times 10^{-4}$ & 64 \\
\cline{2-5}
 & \multirow[t]{4}{*}{MATH} & Full-Tune & $5 \times 10^{-6}$ & 64 \\
 &  & Steering & $5 \times 10^{-4}$ & 64 \\
 &  & LoRA-1 & $1 \times 10^{-4}$ & 64 \\
 &  & LoRA-4 & $1 \times 10^{-4}$ & 64 \\
\cline{1-5} \cline{2-5}
\multirow[t]{8}{*}{Llama-3.1-8B-Instruct} & \multirow[t]{4}{*}{GSM8K} & Full-Tune & $1 \times 10^{-6}$ & 16 \\
 &  & Steering & $2 \times 10^{-4}$ & 16 \\
 &  & LoRA-1 & $6 \times 10^{-4}$ & 16 \\
 &  & LoRA-4 & $1 \times 10^{-4}$ & 16 \\
\cline{2-5}
 & \multirow[t]{4}{*}{MATH} & Full-Tune & $1 \times 10^{-6}$ & 16 \\
 &  & Steering & $2 \times 10^{-4}$ & 16 \\
 &  & LoRA-1 & $3 \times 10^{-4}$ & 16 \\
 &  & LoRA-4 & $3 \times 10^{-4}$ & 16 \\
\cline{1-5} \cline{2-5}
\bottomrule
\end{tabular}